\documentclass[letterpaper, 10 pt, conference]{ieeeconf}  

\IEEEoverridecommandlockouts                              

\usepackage[hidelinks]{hyperref}
\usepackage{subfig}
\usepackage{todonotes}
\usepackage[]{changes}
\usepackage{booktabs}
\usepackage[T1]{fontenc}
\usepackage{siunitx}
\usepackage[nolist]{acronym}
\usepackage{amsmath}
\usepackage{amssymb}
\usepackage{booktabs}
\usepackage{todonotes}
\usepackage[super]{nth}
\usepackage{url}
\title{\LARGE \bf
Destination Prediction Based on Partial Trajectory Data
}
\author{Patrick Ebel$^{1}$, Ibrahim Emre G\"ol$^{1}$, Christoph Lingenfelder$^{2}$ and Andreas Vogelsang$^{1}$
	\thanks{$^{1}$ Technische Universit\"at Berlin, Germany
		{\tt\small [patrick.ebel, andreas.vogelsang]@tu-berlin.de, ibrahim.e.goel@campus.tu-berlin.de}}%
	\thanks{$^{2}$ Mercedes-Benz Innovation Lab, MBition GmbH, Berlin, Germany
		{\tt\small christoph.lingenfelder@daimler.com}}%
}

\makeatletter
\def\ps@IEEEtitlepagestyle{%
	\def\@oddfoot{\mycopyrightnotice}%
	\def\@evenfoot{}%
}
\def\mycopyrightnotice{%
	{\begin{minipage}{\textwidth}
			\footnotesize \copyright 2020 IEEE. Personal use of this material is permitted. Permission from IEEE must be obtained for all other uses, in any current or future media, including reprinting\slash republishing this material for advertising or promotional purposes, creating new collective works, for resale or redistribution to servers or lists, or reuse of any copyrighted component of this work in other works.
		\end{minipage}
	}
	\gdef\mycopyrightnotice{}
}
\makeatother

\begin{document}

\maketitle
\pubidadjcol
\thispagestyle{IEEEtitlepagestyle}
\pagestyle{empty}

\begin{acronym}[ECML-PKDDLL]
	\acro{ADAS}{Advanced Driver Assistance System}
	\acro{ANN}{Artificial Neural Network}
	\acro{API}{Application Programming Interface}
	\acro{BASD}{Bidirectional Space Division Method}
	\acro{CRAWDAD}{A Community Resource for Archiving Wireless Data At Dartmouth}
	\acro{ECML-PKDD}{European Conference on Machine Learning Principles and Practice of Knowledge Discovery in Databases}
	\acro{GPS}{Global Positioning System}
	\acro{GPU}{Graphics Processing Unit}
	\acro{HMM}{Hidden Markov Model}
	\acro{ID}{Identifier}
	\acro{LSTM}{Long Short-Term Memory}
	\acro{MLP}{Multi-Layer Perceptron}
	\acro{MM}{Markov Model}
	\acro{NLP}{Natural Language Processing}
	\acro{POI}{Point of Interest}
	\acro{PPM}{Prediction by Partial Matching}
	\acro{RNN}{Recurrent Neural Network}
	\acro{ST-RNN}{Spatial Temporal Recurrent Neural Network}
\end{acronym}

\hyphenation{time-stam-ped}

\begin{abstract}
	Two-thirds of the people who buy a new car prefer to use a substitute instead of the built-in navigation system. However, for many applications, knowledge about a user's intended destination and route is crucial. For example, suggestions for available parking spots close to the destination can be made or ride-sharing opportunities along the route are facilitated. Our approach predicts probable destinations and routes of a vehicle, based on the most recent partial trajectory and additional contextual data. The approach follows a three-step procedure: First, a \boldmath{$k$}-d tree-based space discretization is performed, mapping GPS locations to discrete regions. Secondly, a recurrent neural network is trained to predict the destination based on partial sequences of trajectories. The neural network produces destination scores, signifying the probability of each region being the destination. Finally, the routes to the most probable destinations are calculated. To evaluate the method, we compare multiple neural architectures and present the experimental results of the destination prediction. The experiments are based on two public datasets of non-personalized, timestamped GPS locations of taxi trips. The best performing models were able to predict the destination of a vehicle with a mean error of \SI{1.3}{\kilo\metre} and \SI{1.43}{\kilo\metre} respectively.
\end{abstract}

\section{Introduction} \label{ch:Introduction}
Knowledge about a user's intended route and destination provides many opportunities to improve the driving experience. A smart travel assistance system that knows in advance where the user is likely to go may offer the following functionalities:

\begin{itemize}
	\item Make smart suggestions for gas or charging stations.
	\item Advise the user to change the planned route in order to avoid a congested road or area.
	\item Suggest Points of Interest (POI).
	\item Show possible parking spots close to the destination.
\end{itemize}

If the driver is using the built-in navigation system, the planned route and destination are known. In such a case the previously named tasks can be solved in a straightforward manner. However, referring to a study~\cite{ValdesDapena.2016} taken by J.D.~Power with more than 13,000 consumers, two-thirds of the people who buy a new car prefer to use a substitute instead of the built-in navigation system. In those cases, the planned route and destination are unavailable for the in-car systems, which makes the route and destination prediction necessary to enable the above-listed features.

\begin{figure}[bt]
	\centering	
	\includegraphics[width=\linewidth]{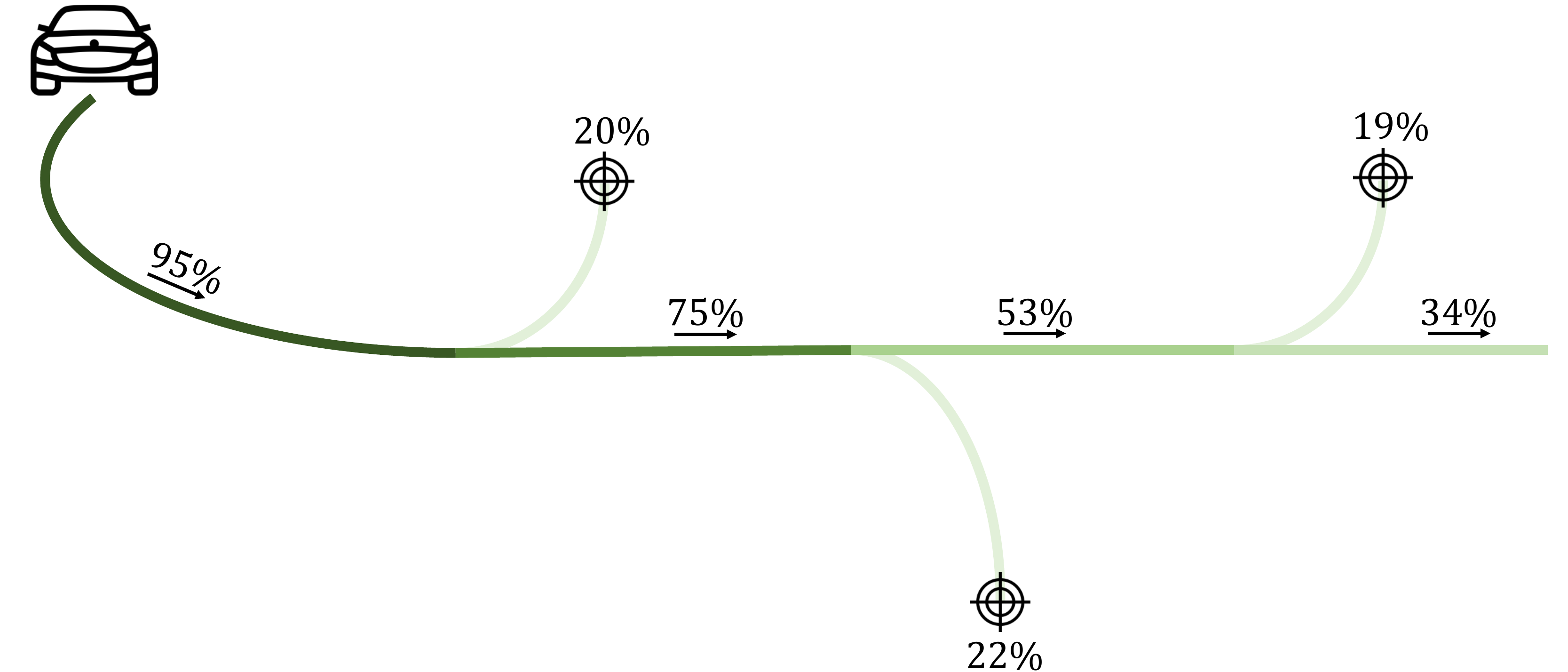}
	\caption{Use Case: Route and Destination Prediction.}
	\label{fig:UseCase}
\end{figure}

In this paper, we propose a \ac{LSTM}-based model using a $k$-d tree-based space partitioning method to predict the future destination and route of a car. To perform a prediction, the model requires a partial trajectory in the form of a sequence of \ac{GPS} coordinates with associated timestamps. The prediction follows a three-step procedure: In the first step, a $k$-d tree-based space discretization is performed, transforming the analyzed area into a set of discrete regions. Thus, each trip is not represented as a sequence of \ac{GPS} locations but as a sequence of regions. This sequence is then, along with additionally retrieved metadata, fed to the \ac{LSTM}. The neural network outputs destination scores, signifying the probability for each region being the destination of the partial trajectory. Subsequently, the highest-scoring, i.e. most probable, routes for the near future can be estimated (see Fig.~\ref{fig:UseCase}). Our approach does not require any personalized data and is, therefore, able to make a forecast based on data collected from a crowd of anonymous users with no knowledge about personal patterns or regularities. The model is evaluated on two datasets containing trajectory information of taxis in two different cities, namely Porto and San Francisco. Additionally, the model is evaluated on the test set provided in the ECML/PKDD 2015 Kaggle Challenge~\cite{KaggleInc..2015} obtaining a score that would have ranked first out of 381 submissions.

\section{Related Work}\label{ch:RelWork}

The recent approaches in the field of destination prediction can be divided into \emph{personalized destination prediction} and \emph{generic destination prediction}. The former approaches~\cite{Krumm.2006,DantasNobreNeto.2018,Manasseh.2013,Stegmann.2018} try to model the mobility patterns of a specific user. Due to the high degree of regularity in personal mobility patterns, they mostly accomplish a very high prediction accuracy. However, the necessary datasets are very limited in their volume and availability. Additionally, recording and saving trajectories directly connected to a person is highly critical due to privacy concerns. It also limits usefulness, because it will fail exactly when a user needs advice most, namely when doing something new.

This work focuses on generic destination prediction where the data is collected from a crowd of anonymous drivers without any knowledge of personal patterns or regularities. The majority of these approaches divide the space into discrete subsets in the first step. This is necessary because the available datasets are too sparse to cover a sufficient amount of query routes~\cite{Wang.2017}. There are mainly two different strategies applied, \emph{road network mapping} and \emph{spatial partitioning}.

In road network mapping, space gets discretized based on an underlying road network. The \ac{GPS} locations are mapped to road segments with a unique identifier. The main drawback of these approaches~\cite{DantasNobreNeto.2018,Lassoued.2017,Li.2016,Simmons.2006} is the high amount of road links, which increases the data sparsity problem.

Spatial partitioning is the process of dividing space into multiple, non-overlapping regions. The spatial partitioning itself can be divided into two approaches, \textit{space-based partitioning} and \textit{trajectory-based partitioning}. The approaches using space-based partitioning~\cite{Krumm.2006,Manasseh.2013,Endo.2017,Pecher.2016} map the locations to an overlaying uniform grid, dividing the space into a set of congruent cells. This has its advantages in its simplicity, fast implementation, and intuitive understanding. The essential downside is that the spatial distribution of trajectories is not considered. Hence, the distribution of data points across the cells or regions might be very imbalanced which can lead to a loss in prediction accuracy~\cite{Xue.2015}. To solve this problem, Xue et. al~\cite{Xue.2015} propose a quantile-based as well as a $k$-d tree-based partitioning strategy, that divide the space based on the density of data points. Due to the more uniform distribution of data points across the grid cells, both result in higher prediction accuracy. Wang et al. introduce a method~\cite{Wang.2017} where space is first divided into uniform grid cells in order to synthesize the nearest cells in the second step. Thus, discrete regions of variable shape are generated.

Besides the different spatial partitioning methods, multiple machine learning approaches are applied in recent work. Most use probabilistic models and especially \acp{MM}, where each state represents a location~\cite{Lassoued.2017, Li.2016, Simmons.2006, Xue.2015}. Thus, each trajectory is modeled as a series of state transitions. Knowing the transitions and their probabilities, the probability of reaching a certain location is calculated. To avoid the data sparsity problem most \acp{MM} are based on low-order Markov processes and, therefore, only incorporate the latest time-steps which limits their capability to model long-term dependencies.

In contrast to probabilistic models, recent efforts have been made using \acp{ANN} for destination prediction. Endo et al.~\cite{Endo.2017} propose an \ac{LSTM}, computing transition probabilities between the grid cells for the next time step is applied. To estimate destination probabilities for destinations further in the future, the authors apply the Monte Carlo principle. The main advantages of \ac{LSTM}-based models lie in their capability to model long-term dependencies and to overcome the data sparsity problem. Liu et al.~\cite{Liu.2016} propose an approach called \ac{ST-RNN}, incorporating distance-specific transition matrices to model geographical dependencies. Br{\'e}bisson et al.~\cite{Brebisson.2015} propose a \ac{MLP}-based solution that won the Kaggle challenge on taxi destination prediction~\cite{KaggleInc..2015}. A \ac{LSTM}-based method which predicts the destination solely based on the individual pick-up and drop-off points of the taxi drivers is presented by Rossi et al.~\cite{Rossi.2019}.

Another solution to the problem are matching-based algorithms. The main idea is to match a query trajectory with recorded trajectories. The predicted destination or route is then equal to the destination or route of the recorded trajectory having the highest similarity to the query trajectory. In the work of Lam et al.~\cite{Lam.2015} and the early work of Froehlich and Krumm~\cite{Froehlich.2008}, trip matching approaches are applied. A hybrid model that combines \acp{MM} with \acp{PPM} is proposed by Dantas Nobre Neto et al.~\cite{DantasNobreNeto.2018, DantasNobreNeto.2016}. However, the main drawback of all matching-based methods is that only routes and destinations which exist in the historical data can be predicted.

\section{Proposed Approach} \label{ch:Approach}
This work combines a trajectory-based space partitioning with an \ac{LSTM}-based multi-input model destination prediction model. 

\subsection{Definitions and Problem Statement}

\textit{Definition 1:} Let $T$ be the set of trajectories $t$. A trajectory $t$ is defined as a sequence $(p_n)_{n=1}^N$, where $p_n \in \mathbb{R}^2$ is a single observation and $N$ denotes the length of the trajectory. Each observation $p_n = (\phi, \lambda)_n$ is composed of its latitudinal $\phi$ and longitudinal $\lambda$ \ac{GPS} coordinates. A trajectory's destination is defined as $y_\text{GPS} = p_N$.

\textit{Definition 2:} A partial trajectory $t_{p} = (p_n)_{n=1}^{N_p}$, is defined as a sub-trajectory of $t$, where $N_p$ is a random variable following the discrete uniform distribution over $\mathbb{N}\cap[2,N-1]$.

\textit{Definition 3:} The haversine distance $D_{h}(p_\text{A},p_\text{B})$ measures the distance between two locations $p_\text{A},p_\text{B} \in \mathbb{R}^2$ on a sphere and is computed as follows:

\begin{equation}\label{eq:Haversine2}
\centering
D_{h}(p_\text{A},p_\text{B})=2 \cdot r \cdot \arctan \left( \sqrt{ \frac{a}{1-a}} \right)\,,\text{with}
\end{equation}

\begin{displaymath}\label{eq:Haversine1}
\centering
a=\sin^2 \left( \frac{\phi_B-\phi_A}{2} \right)+\cos \left( \phi_\text{A} \right)\cos \left(\phi_\text{B} \right)\sin^2 \left(\frac{\lambda_\text{B}-\lambda_\text{A}}{2} \right)\,,
\end{displaymath}

where $r$ is the earth's radius.

\textit{Problem:} We define the destination prediction task as the problem of predicting $y_\text{GPS}$ given a partial trajectory $t_p$.

\subsection{Space Discretization} \label{ch:SpaceDiscretization}
To overcome the data sparsity problem, a $k$-d tree-based partitioning approach is applied. A $k$-d tree is a binary tree in which each node represents a coordinate space in dimension $k$. Each non-leaf node divides the data space of its parent node into two subspaces of equal size. Thus, a $k$-d tree can be used to recursively divide the data space into partitions until a defined number of data points per leaf $n_\text{ppr}$ is reached. In our case, with $k=2$, each coordinate represents a location and each partition represents a region. The number of resulting regions $n_\text{r}$ and their size too, are dependent on $n_\text{ppr}$. For both datasets $n_\text{ppr}$ was heuristically defined such that the regions are small enough in order to achieve a satisfactory accuracy but big enough to avoid data sparsity. 

\begin{figure}
	\subfloat[Denstity map of data points\label{subfig:DensityMapPorto}]{%
		\includegraphics[width=0.48\linewidth]{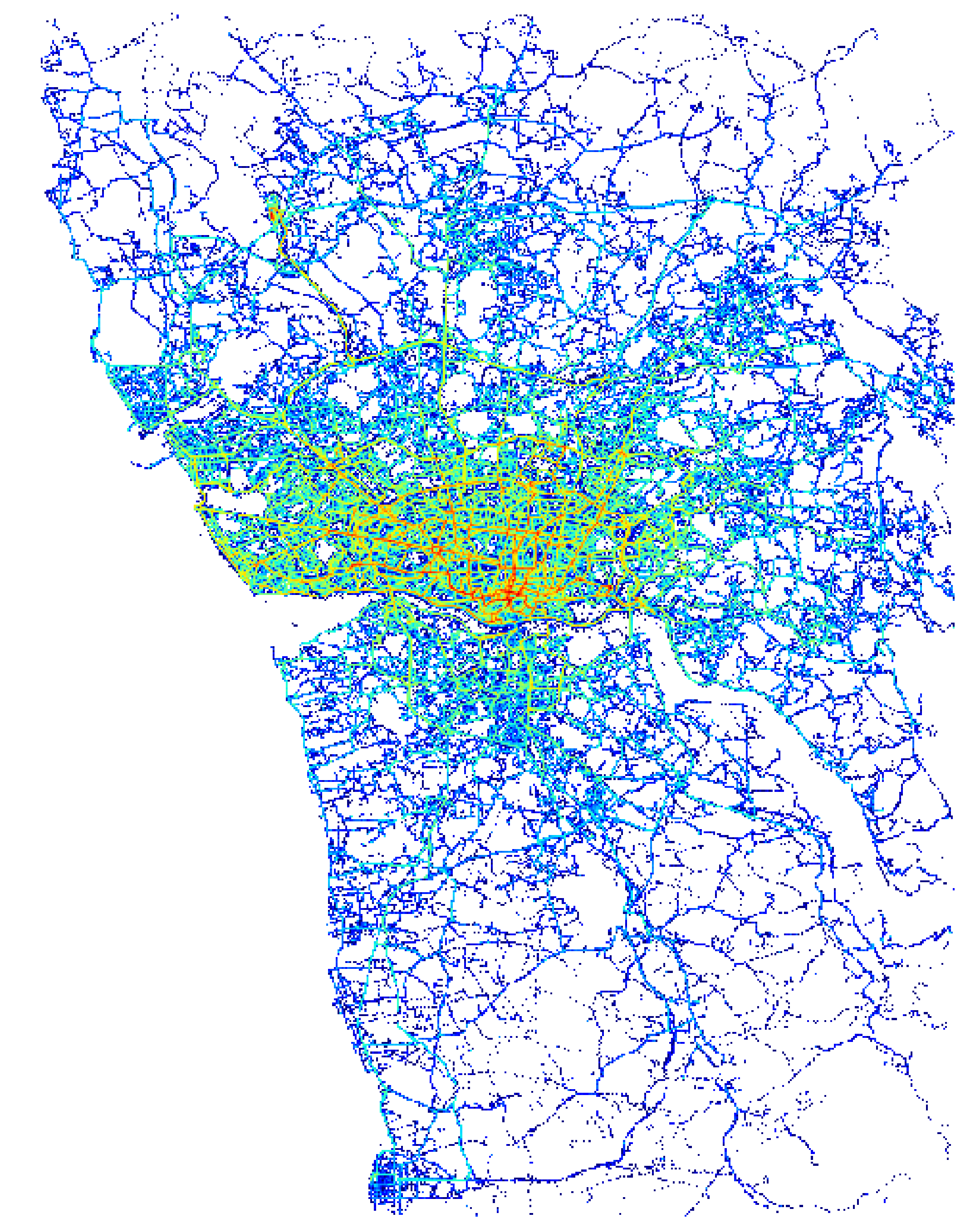}
	}
	\hfill
	\subfloat[Discrete regions\label{subfig:RegionsPorto}]{%
		\includegraphics[width=0.48\linewidth]{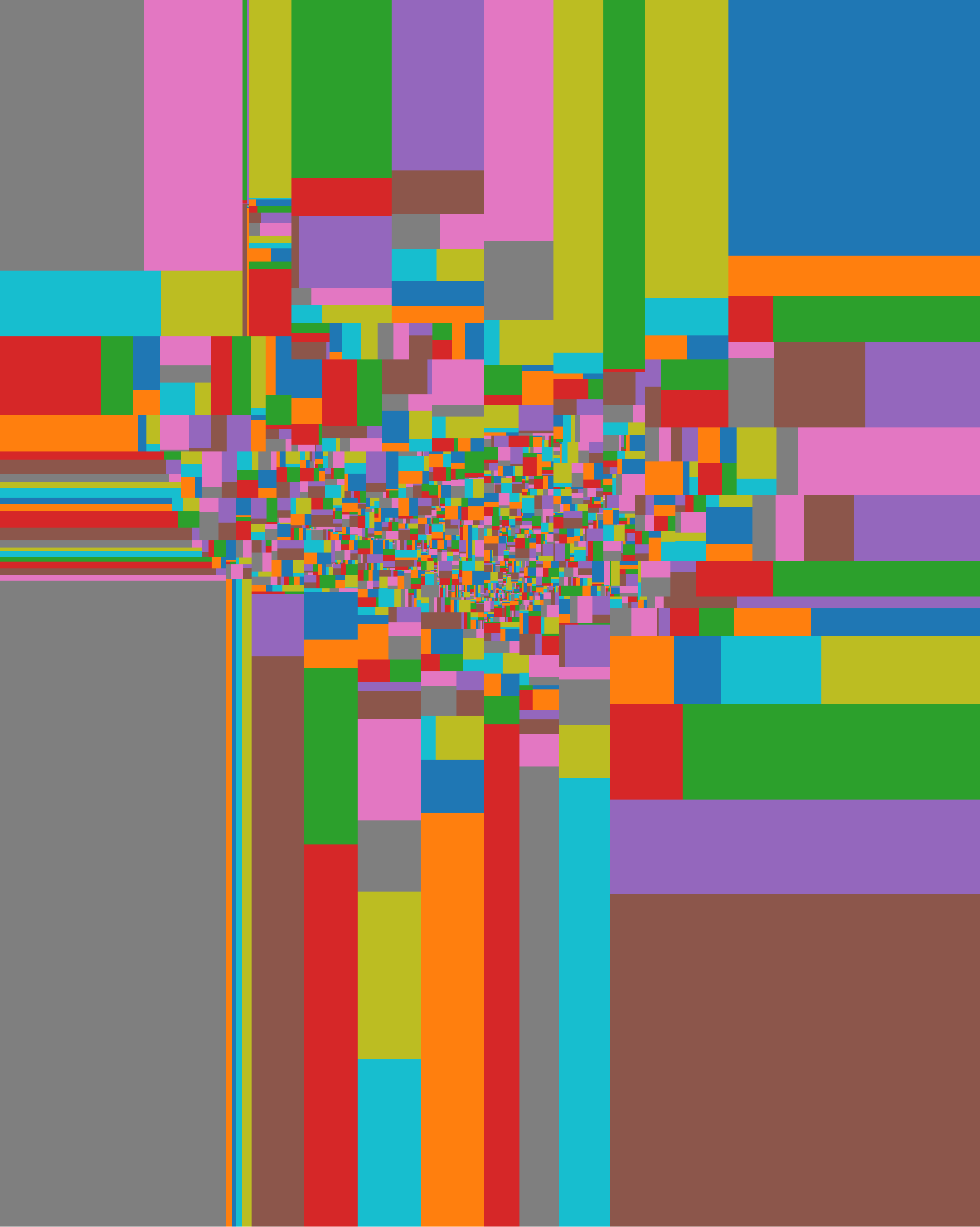}
	}
	\caption{Spatial partitioning of the Porto area.}
	\label{fig:SpatialPartitioning}
\end{figure}

After the discretization each latitude/longitude pair is converted to an integer value $i_\text{r} = 1,\dots,n_\text{r}$, signifying the region which the point is located in. The final partitioning of the Porto dataset is shown in Fig.~\ref{fig:SpatialPartitioning}. We can observe that the regions are smaller in areas with a high density of data points, e.g. the city center or the airport area and larger in more remote areas where less data was collected.

\subsection{Modeling} \label{ch:Modeling}

Due to their ability to keep a memory of previous inputs, \acp{LSTM} are considered to be efficient for time-series prediction. Their main advantage to model long-term dependencies is especially important for the destination prediction task at hand. In contrast to matching-based algorithms, \acp{LSTM} are able to overcome the data sparsity problem, due to their ability to generalize.

\begin{figure}
	\centering
	\includegraphics[width=\linewidth]{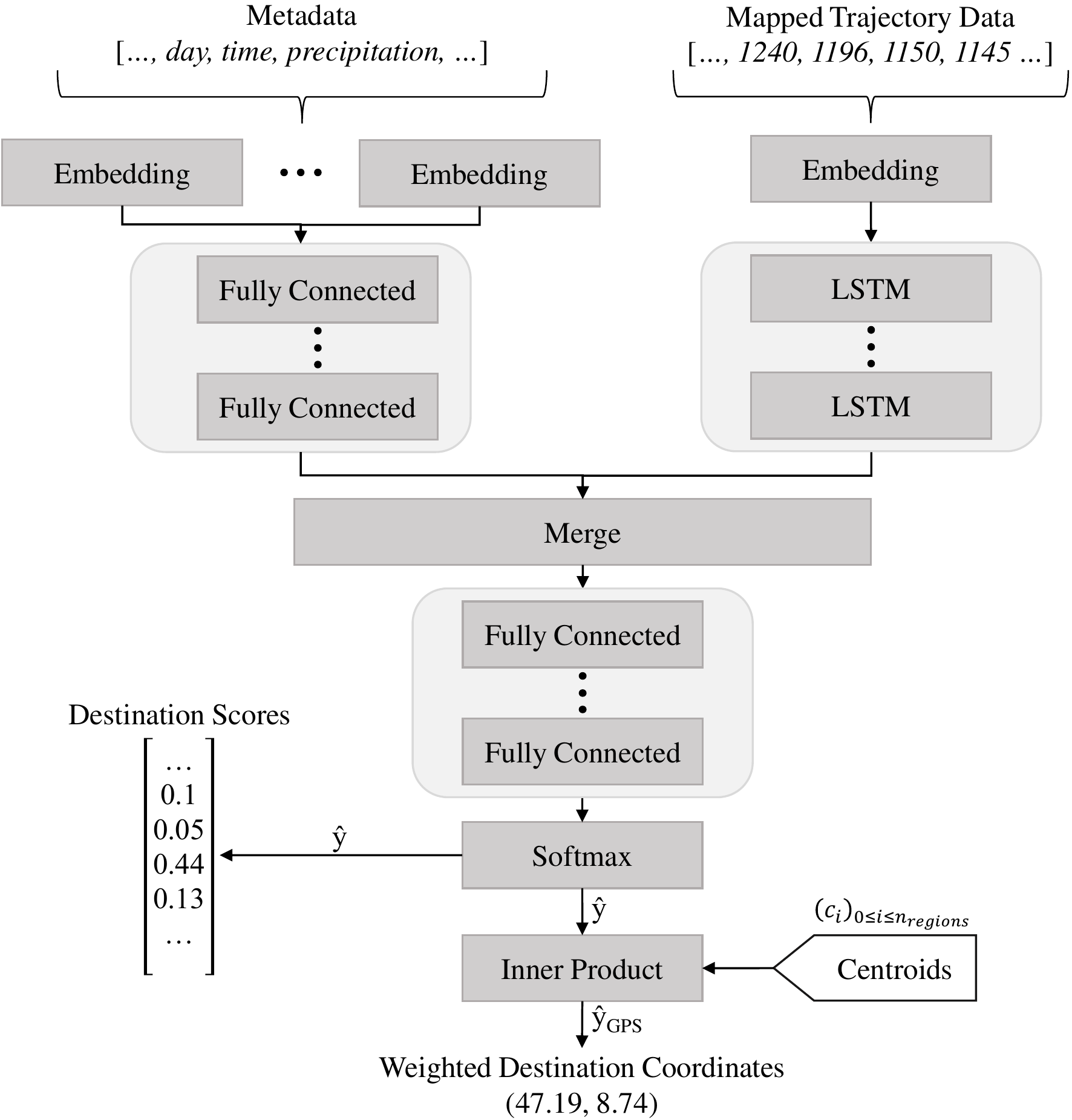}
	\caption[Architecture of the multi-input model.]{Architecture of the \ac{LSTM}-based multi-input model.}
	\label{fig:LSTMModelArchitectureMultiNew}
\end{figure}

In addition to the trajectory information, our approach also processes contextual information, namely the time of the day, the day of the week, the temperature and the precipitation\footnote{We retrieved the weather-related data for San Francisco from \url{www.frontierweather.com} and for Porto from  \url{www.meteoblue.com}}, which are assumed to be constant for each trip. To process the constant contextual information as well as time-series data, namely the trajectory data, the model needs to process the two inputs separately. The architecture of the multi-input model is shown in Fig.~\ref{fig:LSTMModelArchitectureMultiNew}.

The mapped trajectory data is fed to an \emph{Embedding layer} which turns integer values into dense vectors of a fixed size. Thus, each index and therefore each region identifier $i_r$ is mapped to a vector of size $s_\text{embedTrip}$. These vectors are initialized randomly but since the so-called embedding table is part of the model parameters they are tuned during training. The embedding table holds all vectors describing the regions and thus is of size $s_\text{embedTrip} \times n_\text{r}$. The idea of using embeddings to represent integer values was inspired by \ac{NLP}~\cite{Bengio.2003}. The intention to embed the regions is to learn their spatial information and their relationship with each other. The output of the embedding layer is a matrix describing the query trajectory as a sequence of embedding vectors. This matrix is then processed by a stack of $n_{\text{LSTM}}$ many-to-one \ac{LSTM}-layers.

Similar to the trajectory, each attribute of contextual information is mapped to an embedding vector of size $s_\text{embedMeta}$. Afterward, the metadata embeddings are fed into a stack of $n_{\text{denseMeta}}$ fully connected layers. The output of this \ac{MLP} stack is concatenated with the output of the \ac{LSTM} stack, and in turn, fed to a stack of $n_{\text{dense}}$ fully connected layers. The next layer is a \emph{softmax} layer which normalizes its output inputs such that all outputs add up to $1$. Each entry $\hat{y}_i$ of the output vector $\hat{y}$ represents the destination score for region $i$. Thus, $\hat{y}$ can be interpreted as the probability distribution over all regions.

Additionally, the inner product of $\hat{y}$ and $(c_i)_{i=1}^{n_r}$, where $c_i = (\phi^c_i, \lambda^c_i)$ represents the centroid coordinates of region $i$, is computed as follows:

\begin{displaymath}
\hat{y}_\text{GPS} = \begin{pmatrix}
\phi\\
\lambda
\end{pmatrix}_\text{pred} = \begin{pmatrix}
\sum_{i=0}^{n_\text{r}}\hat{y}_i\, \phi^c_i\\
\sum_{i=0}^{n_\text{r}}\hat{y}_i\, \lambda^c_i
\end{pmatrix},
\end{displaymath}

and represents a weighted destination prediction. 

\subsection{Test Design} \label{ch:TestDesign}
Having two outputs $\hat{y}_\text{GPS}$ and $\hat{y}$, the model can be optimized to serve different purposes. Optimizing the model solely regarding $\hat{y}_\text{GPS}$ leads to a superior top-1 destination prediction as it is required in the mentioned Kaggle competition. However, for many intelligent travel assistance systems, a probability distribution over multiple destinations is required. In this case, an optimization regarding $\hat{y}$ is needed.
To measure the performance and optimize the model regarding the two purposes, two error measures are specified. As in most similar works~\cite{Brebisson.2015,Lam.2015,Rossi.2019,Besse.2017} the \emph{Mean Haversine Distance} serves as the basis for both error measures. Inspired by Besse et al.~\cite{Besse.2017}, the first error measure $E_\text{pred}^1(\theta)$ is defined as the mean of the haversine distance between the location of the true destination $y_\text{GPS}$ of the trajectory $t$ and the location of the weighted prediction $\hat{y}_\text{GPS}(t_p, \theta)$, where $\theta$ is the set of model parameters, adjusted during training. Thus, $E_\text{pred}^1(\theta)$ reads as follows:

\begin{equation}\label{eq:Qualitycriterion}
\centering
E_\text{pred}^1(\theta) = \frac{1}{|T|} \sum_{t_p \in T} D_{h}\left(\hat{y}_\text{GPS}\left(t_p,\theta\right), y_\text{GPS}\right)\,.
\end{equation}

However, the optimization of $\hat{y}_\text{GPS}$ using~Eq.\ref{eq:Qualitycriterion} does not imply an optimization of $\hat{y}$. For illustrating the potential flaw, we assume that $y_\text{GPS}$ is equal to the location of region A's centroid $c_A$ and that $c_A$ is the midpoint between $c_B$ and $c_C$ being the centroids of region B and C. During training, the penalty for assigning a score of 0.5 to regions B and C would be zero since $\hat{y}_\text{GPS}$ would be exactly at $c_A$. This also holds for assigning a score of 1 to region A (which would be the wanted solution). Hence, the network has no reason to shift weight to region A.
Thus, for justifying the weights as probability scores and being able to optimize the model with regards to $\hat{y}$, a second error measure $E_\text{pred}^2(\theta)$ is introduced. In $E_\text{pred}^2(\theta)$ the distance from each regions' centroid $c_i$ to the true destination $y_\text{GPS}$ is calculated and weighted based on $\hat{y}_i$:
\begin{equation}\label{eq:Qualitycriterion2}
\centering
E_\text{pred}^2(\theta) = \frac{1}{|T|} \sum_{t_p \in T} \sum_{i=0}^{n_\text{r}} \hat{y}_i \left(t_p,\theta\right) D_{h}\left(c_i, y_\text{GPS}\right)\,.
\end{equation}

In the following, the models are evaluated against $E_\text{pred}^1$ as well as against $E_\text{pred}^2$ and optimized using a weighted combination $E_\text{pred}(\theta)$:

\begin{equation}\label{eq:Qualitycriterion3}
\centering
E_\text{pred}(\theta) = \alpha E_\text{pred}^1(\theta) + (1-\alpha) E_\text{pred}^2(\theta)\,,
\end{equation}

with $0\leq\alpha\leq 1$ being a hyperparameter which controls the importance of the error measures during training.

\subsection{Route Prediction} \label{ch:RoutePred}
The route prediction is based on the assumption that the driver is likely to take the best possible route. Having calculated $\hat{y}$, the destination scores for all regions are known. The route prediction algorithm calculates the route from the last known position to the top-$n$ destinations. Subsequently, each route gets assigned the score of the destination it leads to. Since in most cases, the top-$n$ destinations are in the same area, the calculated routes overlap up to a certain distance. The score of visiting the overlapping parts of the routes is therefore assumed to be equal to the sum of the scores of the individual routes. Therefore, it is possible to retrieve higher scores for near future routes. This is important since, for example, a recommendation system for gas stations will only present a prediction to the driver when the possibility that the driver is taking the assumed route is over some threshold. In that case, it is not necessary to exactly know the final destination but to know the route in the near future.

\section{Evaluation}
In this section, the datasets,  experimental results of the introduced models are presented.

\subsection{Datasets}
Both datasets used in this work are publicly available and contain data collected from taxis. The first dataset, the \textit{Porto dataset}, was published on Kaggle as the basis for a taxi trajectory prediction challenge~\cite{KaggleInc..2015} and is also used in several other works~\cite{Endo.2017,Brebisson.2015,Rossi.2019,Lam.2015,Besse.2017}. It contains 1,710,670 trajectories of 442 taxis operating in the city of Porto, Portugal. The recording of the data took place over a period of one year starting in July 2013. The second dataset~\cite{Piorkowski.2009} contains 927,976 trajectories of 536 taxis collected over 30 days in San Francisco, USA. This dataset was also processed in different other works~\cite{Rossi.2019,Besse.2017} and is further referred to as the \textit{San Francisco dataset}. In contrast to the Porto dataset where the trajectories are given as a univariate time-series with an interval of \SI{15}{\second}, the update interval varies for the observations in the San Francisco dataset with a mean of \SI{63.3}{\second} and a standard deviation of \SI{52.58}{\second}.

\subsection{Data Preprocessing}\label{ch:dataProcessing}
Before the data can be fed to a model it needs to be processed. The data preprocessing consist of four different steps (see \autoref{tab:dataPreprocessingSteps}): (1) Initially, all trips that are either extremely short ($t < 2\,\text{min}$), extremely long ($t > 120\,\text{min}$) or consist of only a single datapoint are deleted. This is done based on the assumption that most of these irregularities occur due to recording issues. (2) To further enhance the data quality, trips containing erroneous data points, e.g. due to \ac{GPS} errors or incorrectly handled taximeters are conditioned as follows: If the assumed speed between two consecutive points exceeds 240 km/h, the outliers are smoothed by applying a moving median filter. (3) Afterward, all trips that still contain locations outside of the defined area (exemplarily displayed for Porto in Fig.~\ref{fig:SpatialPartitioning}) are deleted. (4) Roundtrips or sightseeing trips have no value for destination prediction models but are existent especially in taxi data. To clean those trips we introduce a \emph{roundtrip factor} $\tau$, describing the relation between the length of a trip and the linear distance between start and destination:

\begin{equation}\label{eq:rtf}
\tau=\frac{\sum_{n=1}^{N}D_H(p_n,\, p_{n+1})}{D_H(p_1, p_N)}\,.
\end{equation}

\begin{table}
	\small
	\caption{Data preprocessing steps.}
	\label{tab:dataPreprocessingSteps}
	\centering
	\begin{tabular}{ccc} 
		\toprule
		Step & \multicolumn{2}{c}{Number of trips}\\
		\cmidrule(lr){2-3}
		& Porto & San Francisco\\ 
		\midrule
		-	& 1,710,670 (100.0\,\%) & 927,976 (100.0\,\%)	\\
		(1)	& 1,638,681 (95.79\,\%) & 820,108 (88.37\,\%)\\
		(2)	& 1,638,681 (95.79\,\%) & 820,108 (88.37\,\%)\\
		(3)	& 1,630,112 (95.29\,\%) & 815,403 (87.87\,\%)\\
		(4)	& 1,545,240 (90.33\,\%) & 700,197 (75.44\,\%) \\
		\bottomrule
	\end{tabular}
\end{table}

The city topology, as well as the frequency of the update intervals, affects what can be considered as a roundtrip. Thus, the threshold for $\tau$ needs to be chosen separately for each dataset. For the Porto dataset, we choose $\tau_\text{P}=3.5$, which corresponds to the \nth{95} percentile of the distribution of $\tau_\text{P}$ over all trips in the dataset at this preprocessing step. Thus, all trips that are longer than 3.5 times the beeline between their start and destination are deleted. For the San Francisco dataset, we choose $\tau_\text{SF}=2.65$, being more restrictive, due to the longer update intervals between two consecutive data points. After preprocessing, the Porto dataset is reduced to 1,545,240 trajectories which accounts for 90.33\,\% of the data. The San Francisco dataset is reduced to 700,197 trajectories (75.44\,\%). 

\subsection{Hyperparameter Optimization}
To find the best performing set of parameters\footnote{\url{https://doi.org/10.6084/m9.figshare.11698500}}, a hyperparameter optimization is performed for all models on two NVIDIA Quadro P5000 \acp{GPU}. To evaluate the models and to provide the necessary comparability, all models are trained, evaluated and tested on the same preprocessed datasets. Due to the large amount of data the models are trained on 90\,\% of the data randomly sampled from the datasets. The remaining 10\,\% are equally split in validation and test set which still results in a set size of 35.000 trajectories for the smaller San Francisco dataset. 

During optimization we found that if $\alpha \neq 1$ at the beginning of the training process, $E_\text{pred}^1$ and $E_\text{pred}^2$ are improving very slowly. However, a good method to prevent this behavior and to accelerate the training process is to set $\alpha=1$ for the first $e$ epochs. This leads to a fast reduction of $E_\text{pred}^1$ and an accompanying slower reduction of $E_\text{pred}^2$. If $\alpha$ is decreased after $e$ epochs, $E_\text{pred}^2$ converges close to $E_\text{pred}^1$.

\subsection{Experimental Results}
In addition to the \ac{LSTM}-based approach, three alternative approaches are evaluated. The preprocessing procedure as well as the space discretization process are equal for all the approaches.

The first approach, called \emph{Baseline Algorithm}, is solely based on the trigonometrical relationship between the partial trajectory and the centroid coordinates. First, a set of destination candidates, consisting of the top-$k$ most visited regions is calculated. The predicted destination is then equal to the centroid coordinates of the destination candidate, closest to the extension of the straight line going through the first and last point of the partial trajectory.

The second approach is based on an \ac{MLP}. Due to the architectural requirements of \acp{MLP}, the input has to be a fixed-size vector. As in the work of Brebisson et al.~\cite{Brebisson.2015}, this problem is overcome by feeding the model with the first and last $j$ locations, respectively regions, of the query trajectory.

Additionally, a single-input, \ac{LSTM}-based approach is evaluated, that only takes the query trajectory as input and serves as a reference to quantify whether the prediction accuracy can be improved by considering contextual information.

Table~\ref{tab:resultComparison} shows the results achieved on the test sets. Since the Baseline Algorithm only outputs a single destination prediction, it is only evaluated with respect to $E_\text{pred}^1$.

\begin{table}
	\small
	\caption{Comparison of the different models.}
	\label{tab:resultComparison}
	\centering
		\begin{tabular}{lcccc} 
			\toprule
			 & \multicolumn{2}{c}{Porto} & \multicolumn{2}{c}{San Francisco}\\
			\cmidrule(lr){2-5}
			Model 					& $E_\text{pred}^1$ & $E_\text{pred}^2$		& $E_\text{pred}^1$ 	& $E_\text{pred}^2$\\ 
			\toprule
			Baseline Algorithm		& \SI{2504}{\metre} 	& -					& \SI{2103}{\metre} 	& -\\
			Single-Input \ac{MLP}	& \SI{1595}{\metre} 	& \SI{1635}{\metre}	& \SI{1573}{\metre} 	& \SI{1672}{\metre}\\
			Single-Input \ac{LSTM}	& \SI{1460}{\metre} 	& \SI{1567}{\metre}	& \SI{1300}{\metre} 	& \SI{1462}{\metre}\\
			Multi-Input \ac{LSTM}	& \SI{1430}{\metre} 	& \SI{1480}{\metre}	& \SI{1315}{\metre} 	& \SI{1388}{\metre}\\
			\bottomrule
		\end{tabular}
\end{table}

In general the models behave similarly on both datasets with $E_\text{pred}^2$ being slightly higher than $E_\text{pred}^1$. This originates from the fact, that scores for regions that are on opposite sides from the true destination eliminate each other for some degree when $\hat{y}_\text{GPS}$ is calculated which results in a lower $E_\text{pred}^1$.

Regarding the different approaches, both \ac{LSTM}-based models outperform the \ac{MLP}-based model. This supports the assumption that \ac{LSTM}-based models are superior when it comes to handling long-term dependencies. However, the small difference in performance implies that the first and last $j$ locations of a query trajectory hold most of the information regarding the final destination. As displayed, concerning $E_\text{pred}^2$, the multi-input \ac{LSTM}-based models are superior to the single-input model, which only considers the trajectory information. This strengthens the presumption that the consideration of contextual information contributes to better prediction performance. However, the difference in performance between the two approaches is less than $6\,\%$ for $E_\text{pred}^2$ and for $E_\text{pred}^1$ the single-input model even outperforms the multi-input model on the San Francisco dataset. This may be due to the limitation to taxi data. For private cars, the correlation between the metadata and the destination may be higher, due to regularities introduced by commuting patterns.

\begin{figure}
	\centering	
	\includegraphics[width=\linewidth]{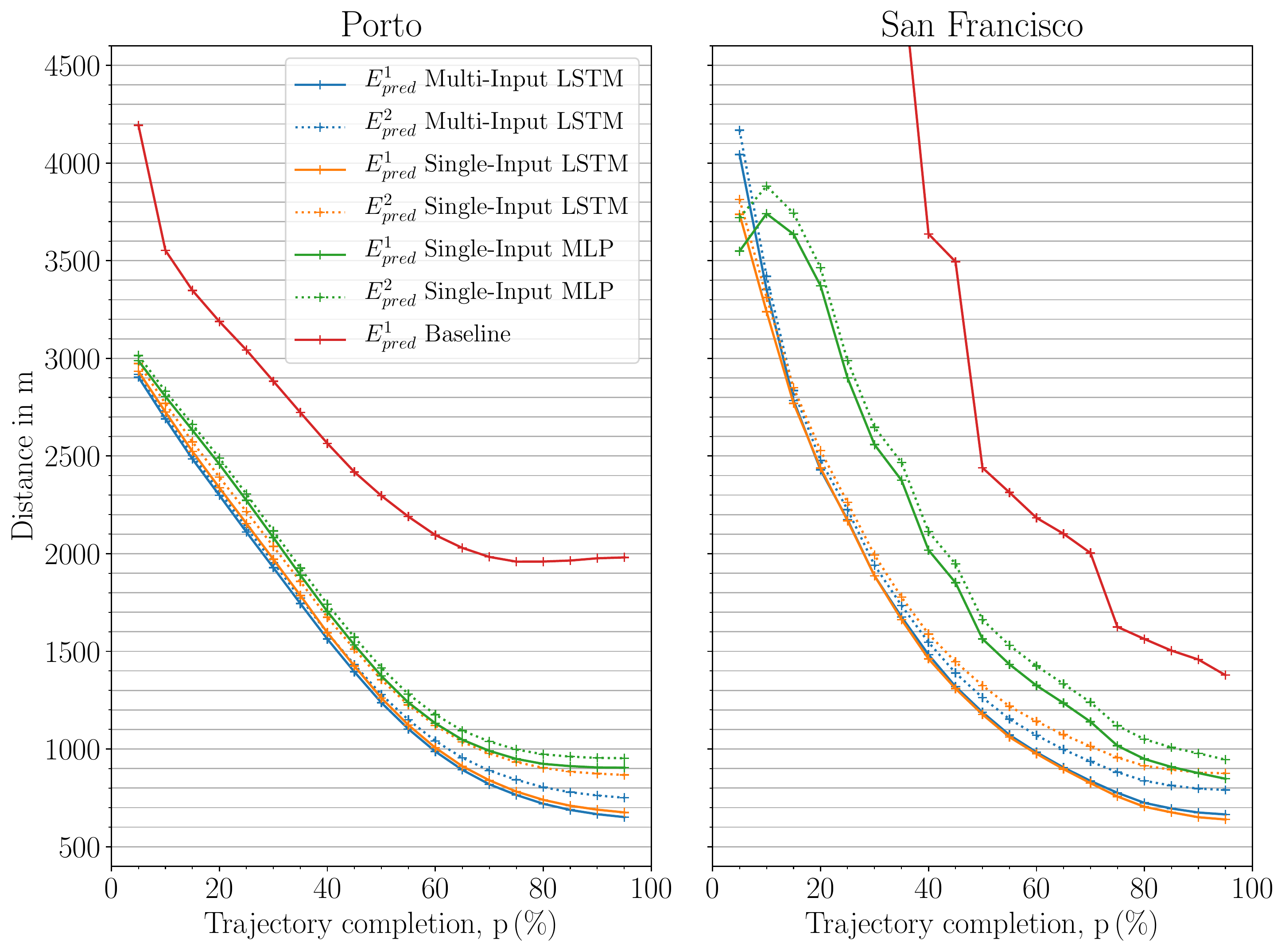}
	\caption{$E_\text{pred}^1$ and $E_\text{pred}^2$ according to trajectory completion.}
	\label{fig:partialScores}
\end{figure}

Fig.~\ref{fig:partialScores} shows $E_\text{pred}^1$ and $E_\text{pred}^2$ of the final three \ac{ANN}-based models and the Baseline Algorithm according to the given proportion $p$ of the full trajectory. If only $5\,\%$ of the trajectory is given, both errors of the \ac{ANN}-based models are between \SI{2.9}{\kilo\metre} and \SI{3}{\kilo\metre} on the Porto dataset and between \SI{3.5}{\kilo\metre} and \SI{4.3}{\kilo\metre} on the San Francsisco dataset. If each query trajectory consists of 50\,\% of its full trajectory, on average, the weighted predictions of the multi-input models are \SI{1.237}{\kilo\metre} (Porto) and \SI{1.188}{\kilo\metre} (San Francisco) away from the true destination. For all approaches, $E_\text{pred}^1$ and $E_\text{pred}^2$ decrease with an increasing length of the given partial trajectory. The gap between $E_\text{pred}^1$ and $E_\text{pred}^2$ increases accordingly for each of the models. However, there is a larger difference for the single-input \ac{LSTM} compared to the multi-input \ac{LSTM}. 

\begin{figure}
	\centering
	\includegraphics[width=\linewidth]{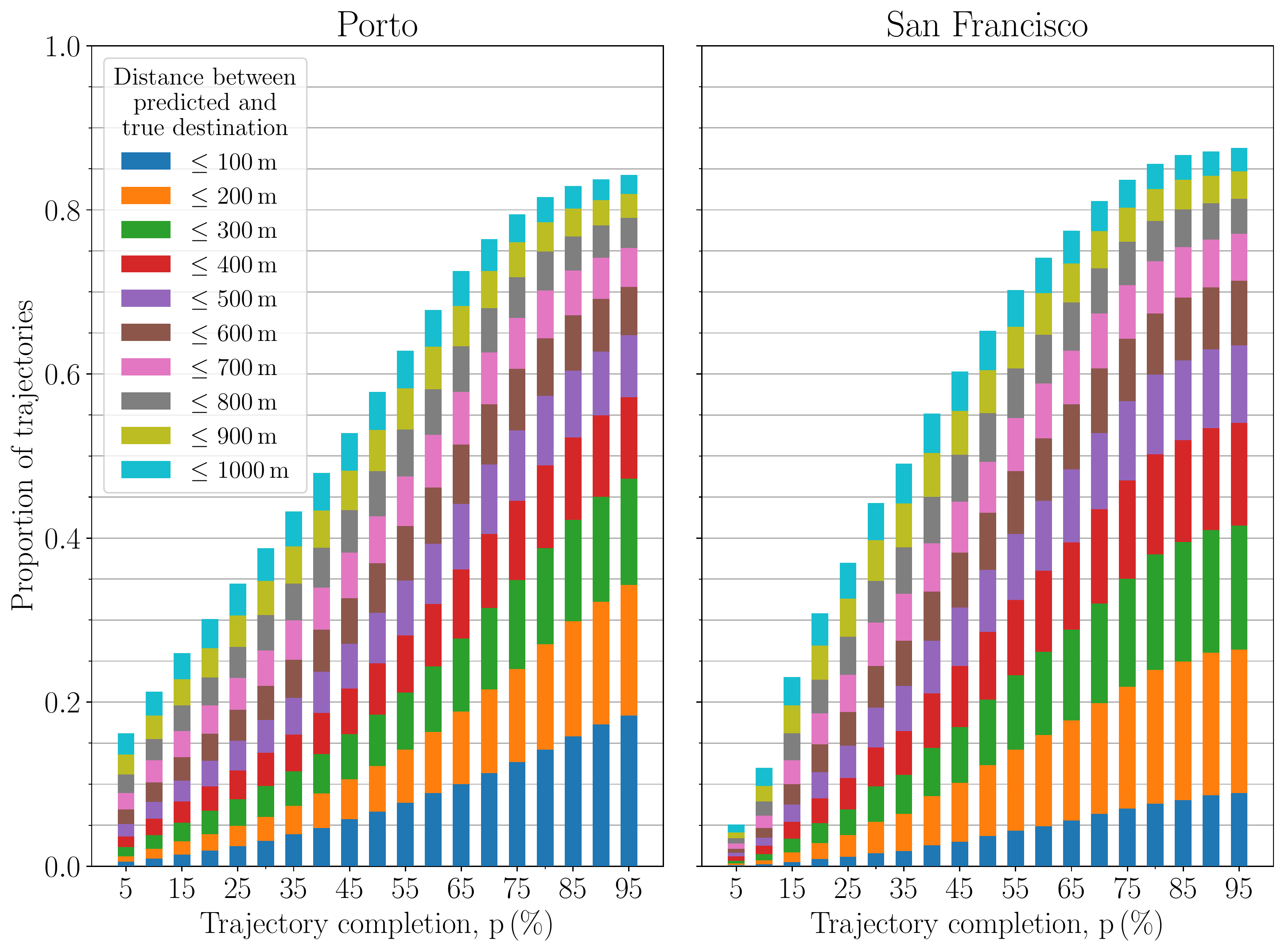}
	\caption{Distribution of $E_\text{pred}^1$ according to trajectory completion.}
	\label{fig:partialScoresDistribution}
\end{figure}

Fig.~\ref{fig:partialScoresDistribution} shows the distribution of $E_\text{pred}^1$ of the final multi-input LSTM-based models according to the trajectory completion. It can be observed that the prediction accuracy improves regularly with the length of the given trajectory. Comparing the datasets, we can observe that, for low completion rates, the predictions are more accurate on the Porto dataset and that the proportion of predictions that are closer than \SI{100}{\metre} to the true destination is twice as high as for the San Francisco dataset.

We additionally evaluated the models based on a snippet of length $t$, which is randomly sampled from the full trajectory is given. If 2 minutes of the trip are given, $E_\text{pred}^1$ and $E_\text{pred}^2$ of the \ac{LSTM}-based models are about \SI{1.5}{\kilo\metre} for both datasets.

\begin{figure}
	\subfloat[$t=\SI{15}{\second}$\label{subfig:TripSequence1}]{%
		\includegraphics[width=0.48\linewidth]{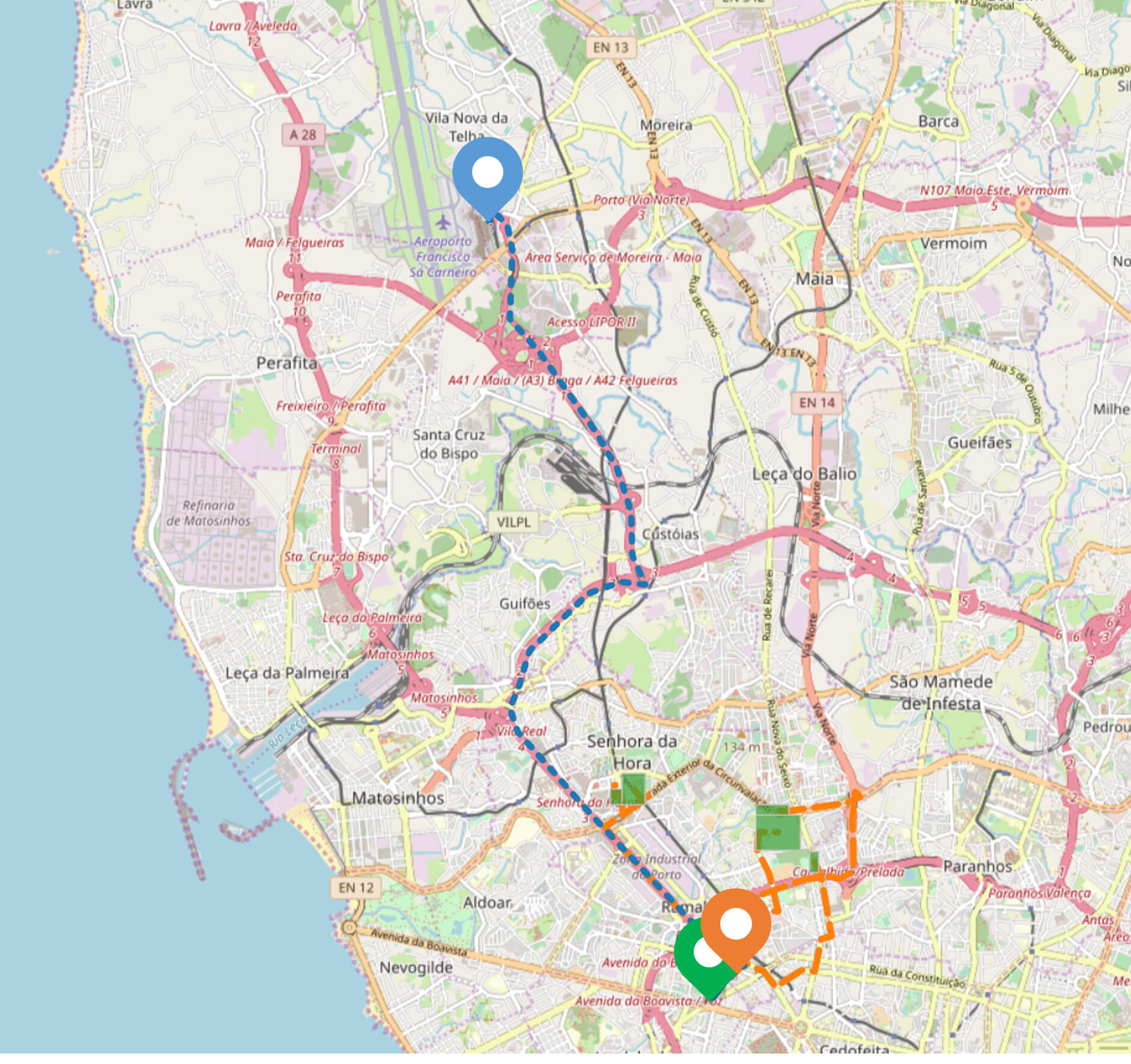}
	}
	\hfill
	\subfloat[$t=\SI{270}{\second}$\label{subfig:TripSequence2}]{%
		\includegraphics[width=0.48\linewidth]{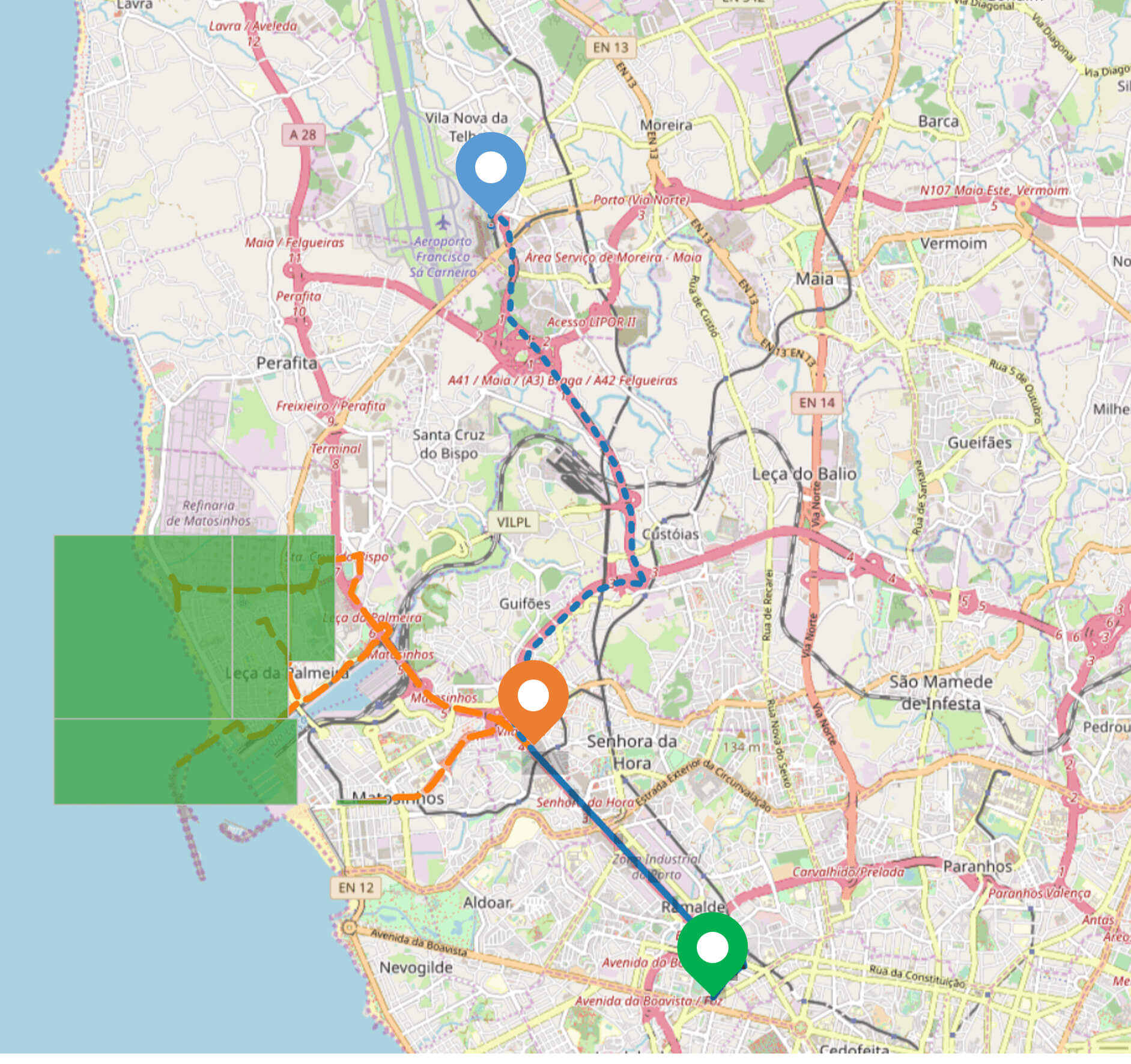}
	}
	\hfill
	\subfloat[$t=\SI{315}{\second}$\label{subfig:TripSequence3}]{%
		\includegraphics[width=0.48\linewidth]{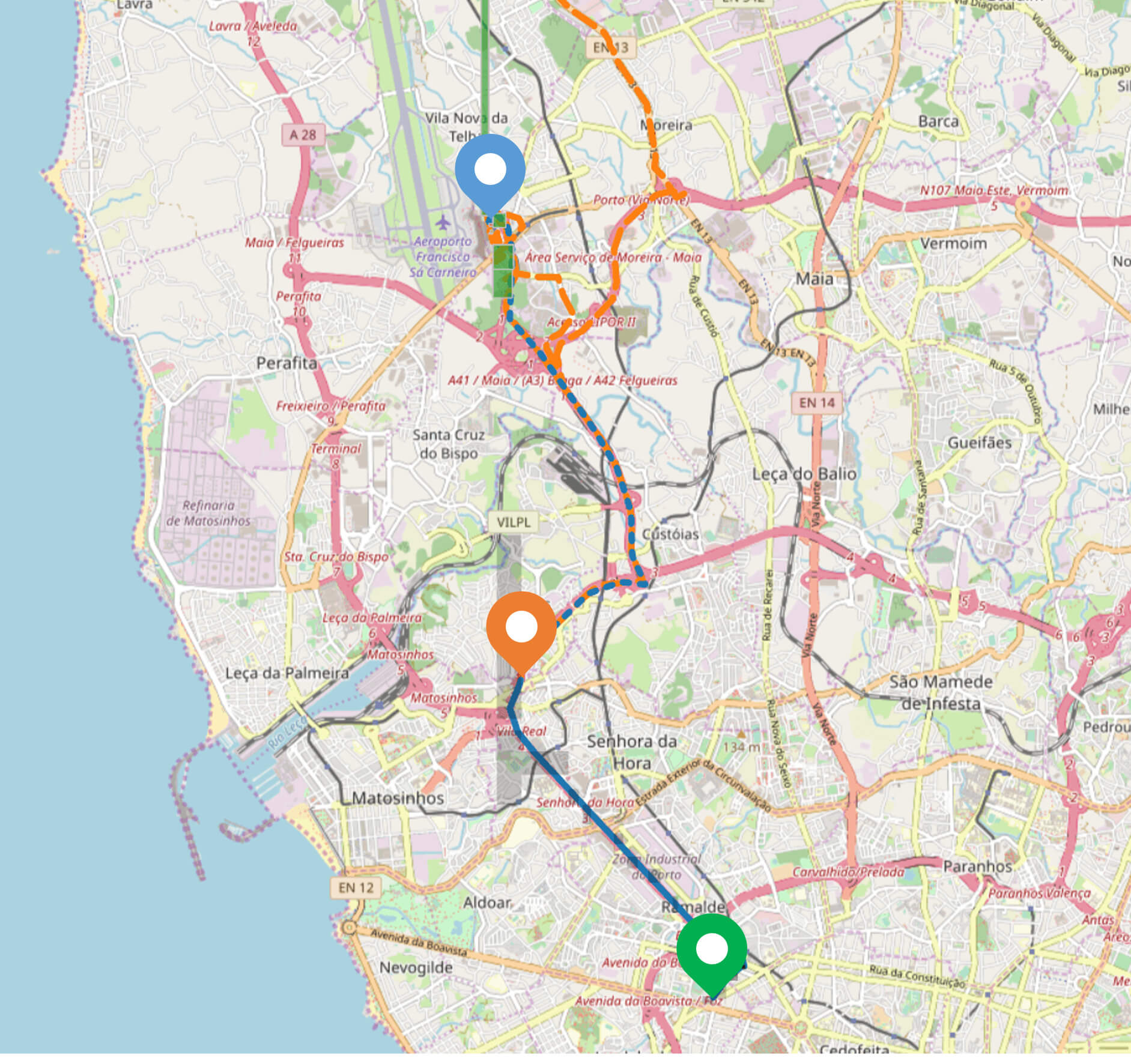}
	}
	\hfill
	\subfloat[$t=\SI{600}{\second}$\label{subfig:TripSequence4}]{%
		\includegraphics[width=0.48\linewidth]{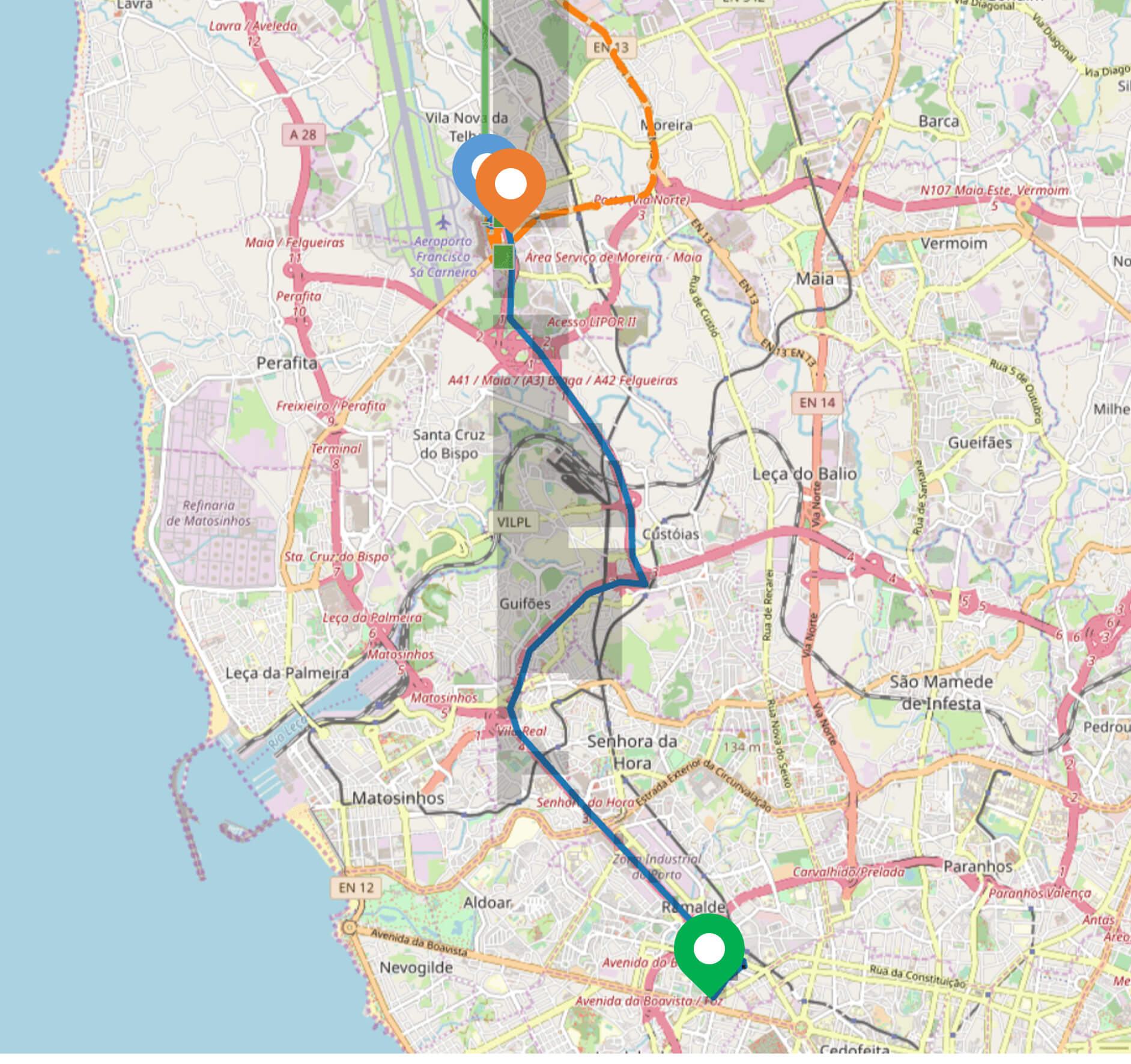}
	}
	\caption{Route and destination prediction.}
	\label{fig:SequentialTrip}
\end{figure}

Fig.~\ref{fig:SequentialTrip} shows a route and destination prediction for a query trajectory, starting in the city center (green marker) and ending at the airport (blue marker). 
In Fig.~\ref{subfig:TripSequence1}, the car has only traveled a small distance (orange marker) and therefore the top-5 predicted regions (green squares) are widely spread. However, the traveling direction of the car is already roughly determined. At the time of the second prediction (Fig.~\ref{subfig:TripSequence2}), made close to the highway exit which leads to the harbor, all predicted regions lie in the harbor area. Shortly after passing the exit (Fig.~\ref{subfig:TripSequence2}), the top-5 predictions jump to the airport area and remain there until the car arrives. Thus, after roughly 50\,\% of the route, the prediction is already very accurate and the top-5 predicted routes (dotted orange lines) match the true ongoing trajectory (dotted blue line) to a high degree. Thus, the approach is not only able to determine that the driver is very likely to go to the airport, but also to predict the route he is going to take.

\subsection{Kaggle}
The Kaggle competition is already over, but it is still possible to submit results and receive a ranking on an unknown test set. Our best multi-input model, optimized only against $E_\text{pred}^1$ achieved a mean haversine distance of \SI{1.995}{\kilo\metre} and would have ranked first out of 381 submissions.

\section{Advantages of the proposed Method}
Our method is designed to predict the route and destination of a car without using personalized location data. In our approach, data is collected from a crowd of anonymous users with no knowledge about personal patterns or regularities. Therefore, no personalized data is needed to train the models or to make predictions. Personalized destination prediction approaches~\cite{Krumm.2006,DantasNobreNeto.2018,Manasseh.2013,Stegmann.2018,Rossi.2019} try to learn the mobility patterns of a specific user to predict future movements. Compared to those approaches, our method is more broadly applicable in practice, especially when it comes to private cars since the handling of personalized location data from customers is highly critical and raises privacy concerns. Additionally, our model can make a prediction even if the user has never been to the city before.

The presented approach not only solves the \mbox{top-1} destination prediction problem but can also predict multiple destinations and their probabilities. The probability assignment is an important advantage when it comes to practical use cases. It allows recommendation systems to consider whether a suggestion based on the destination prediction should be made to the driver or not. This is important since too many inappropriate suggestions based on unsure destination predictions would lead to the user no longer using the respective system.

Further, our approach achieves good prediction accuracies not only for partial trajectories of any length but also for partial trajectories where the starting point is not known. This may be necessary because a constant recording of the trajectory may not be permitted in practice due to privacy regulations.

\section{Conclusions and Future Work}

For many intelligent applications, which aim to improve the driving experience, knowledge about a driver's intended route and destination is crucial. This work introduces three destination prediction models based on \acp{ANN}, able to solve this problem. The chosen datasets are analyzed, cleaned and processed. The location data is transformed using a $k$-d tree-based spatial partitioning approach. The results, achieved on the Porto and San Francisco datasets, show that the best performing models are able to predict the destination based on randomly cut query trajectories with an average accuracy of \SI{1.43}{\kilo\metre} and \SI{1.3}{\kilo\metre} respectively. Even without considering the additionally given contextual data of the Kaggle competition, the multi-input \ac{LSTM}-based model would have scored first out of $381$ approaches. Additionally, the models can predict multiple destinations and their probabilities at any time of the trajectory. In combination with the introduced route prediction, valuable input for multiple in-car applications can be produced.

One potential limitation of our work is that we were only able to evaluate it on data collected from taxis. This may introduce bias compared to the results we would have achieved for private cars. In the next steps, we want to evaluate the method on data collected from private cars and analyze the impact of the metadata used for prediction.

As for most deep-learning approaches, our models lack explainability. For future work, it may be interesting to implement an attention mechanism. The neural attention mechanism has the possibility to enhance the interpretability~\cite{Xu.2015} and would therefore allow us to draw conclusions which parts of the trajectories are especially important for predicting the destination.

\bibliographystyle{IEEEtran}
\bibliography{mybibfile}

\begin{thebibliography}{10}
\providecommand{\url}[1]{#1}
\csname url@rmstyle\endcsname
\providecommand{\newblock}{\relax}
\providecommand{\bibinfo}[2]{#2}
\providecommand\BIBentrySTDinterwordspacing{\spaceskip=0pt\relax}
\providecommand\BIBentryALTinterwordstretchfactor{4}
\providecommand\BIBentryALTinterwordspacing{\spaceskip=\fontdimen2\font plus
\BIBentryALTinterwordstretchfactor\fontdimen3\font minus
  \fontdimen4\font\relax}
\providecommand\BIBforeignlanguage[2]{{%
\expandafter\ifx\csname l@#1\endcsname\relax
\typeout{** WARNING: IEEEtran.bst: No hyphenation pattern has been}%
\typeout{** loaded for the language `#1'. Using the pattern for}%
\typeout{** the default language instead.}%
\else
\language=\csname l@#1\endcsname
\fi
#2}}

\bibitem{ValdesDapena.2016}
\BIBentryALTinterwordspacing
P.~Valdes-Dapena, ``{Most drivers who own cars with built-in GPS systems use
  phones for directions},'' 2016. [Online]. Available:
  \url{https://money.cnn.com/2016/10/10/autos/car-navigation-frustration/index.html}
\BIBentrySTDinterwordspacing

\bibitem{KaggleInc..2015}
\BIBentryALTinterwordspacing
{Kaggle Inc.}, ``{ECML/PKDD 15: Taxi Trajectory Prediction},'' 2015. [Online].
  Available:
  \url{https://www.kaggle.com/c/pkdd-15-predict-taxi-service-trajectory-i/data}
\BIBentrySTDinterwordspacing

\bibitem{Krumm.2006}
J.~Krumm and E.~Horvitz, ``{Predestination: Inferring Destinations from Partial
  Trajectories},'' in \emph{{UbiComp}}, 2006, pp. 243--260.

\bibitem{DantasNobreNeto.2018}
F.~{Dantas Nobre Neto}, C.~d.~S. Baptista, and C.~E.~C. Campelo, ``{Combining
  Markov model and Prediction by Partial Matching compression technique for
  route and destination prediction},'' \emph{{Knowledge-Based Systems}}, vol.
  154, pp. 81--92, 2018.

\bibitem{Manasseh.2013}
C.~Manasseh and R.~Sengupta, ``{Predicting driver destination using machine
  learning techniques},'' in \emph{{16th International IEEE Conference on
  Intelligent Transportation Systems (ITSC 2013)}}, 2013, pp. 142--147.

\bibitem{Stegmann.2018}
R.~A. Stegmann, I.~{\v{Z}}liobait{\.{e}}, T.~Tolvanen, J.~Hollm{\'e}n, and
  J.~Read, ``{A survey of evaluation methods for personal route and destination
  prediction from mobility traces},'' \emph{{Wiley Interdisciplinary Reviews:
  Data Mining and Knowledge Discovery}}, p. e1237, 2018.

\bibitem{Wang.2017}
L.~Wang, M.~Wang, T.~Ku, Y.~Cheng, and X.~Guo, ``{A hybrid model towards moving
  route prediction under data sparsity},'' in \emph{{2017 20th International
  Conference on Information Fusion}}, 2017, pp. 1--8.

\bibitem{Lassoued.2017}
Y.~Lassoued, J.~Monteil, Y.~Gu, G.~Russo, R.~Shorten, and M.~Mevissen, ``{A
  hidden Markov model for route and destination prediction},'' in \emph{{2017
  IEEE 20th International Conference on Intelligent Transportation Systems
  (ITSC)}}, 2017, pp. 1--8.

\bibitem{Li.2016}
X.~Li, M.~Li, Y.-J. Gong, X.-L. Zhang, and J.~Yin, ``{T-DesP: Destination
  Prediction Based on Big Trajectory Data},'' \emph{{IEEE Transactions on
  Intelligent Transportation Systems}}, pp. 2344--2354, 2016.

\bibitem{Simmons.2006}
R.~Simmons, B.~Browning, Y.~Zhang, and V.~Sadekar, ``{Learning to Predict
  Driver Route and Destination Intent},'' in \emph{{2006 IEEE Intelligent
  Transportation Systems Conference}}, 2006, pp. 127--132.

\bibitem{Endo.2017}
Y.~Endo, K.~Nishida, H.~Toda, and H.~Sawada, ``{Predicting Destinations from
  Partial Trajectories Using Recurrent Neural Network},'' in \emph{{Advances in
  Knowledge Discovery and Data Mining}}, 2017, pp. 160--172.

\bibitem{Pecher.2016}
P.~Pecher, M.~Hunter, and R.~Fujimoto, ``{Data-Driven Vehicle Trajectory
  Prediction},'' in \emph{{Proceedings of the 2016 ACM SIGSIM Conference on
  Principles of Advanced Discrete Simulation}}, ser. {SIGSIM-PADS '16}.\hskip
  1em plus 0.5em minus 0.4em\relax New York, NY, USA: ACM, 2016, pp. 13--22.

\bibitem{Xue.2015}
A.~Y. Xue, J.~Qi, X.~Xie, R.~Zhang, J.~Huang, and Y.~Li, ``{Solving the data
  sparsity problem in destination prediction},'' \emph{{The VLDB Journal}},
  vol.~24, no.~2, pp. 219--243, 2015.

\bibitem{Liu.2016}
Q.~Liu, S.~Wu, L.~Wang, and T.~Tan, ``{Predicting the Next Location: A
  Recurrent Model with Spatial and Temporal Contexts},'' in \emph{{Proceedings
  of the Thirtieth AAAI Conference on Artificial Intelligence}}, ser.
  {AAAI'16}.\hskip 1em plus 0.5em minus 0.4em\relax {AAAI Press}, 2016, pp.
  194--200.

\bibitem{Brebisson.2015}
A.~d. Br{\'e}bisson, {\'E}.~Simon, A.~Auvolat, P.~Vincent, and Y.~Bengio,
  ``{Artificial Neural Networks Applied to Taxi Destination Prediction},''
  \emph{{CoRR}}, vol. abs/1508.00021, 2015.

\bibitem{Rossi.2019}
A.~Rossi, G.~Barlacchi, M.~Bianchini, and B.~Lepri, ``{Modelling Taxi Drivers'
  Behaviour for the Next Destination Prediction},'' \emph{{IEEE Transactions on
  Intelligent Transportation Systems}}, pp. 1--10, 2019.

\bibitem{Lam.2015}
H.~T. Lam, E.~Diaz-Aviles, A.~Pascale, Y.~Gkoufas, and B.~Chen, ``{(Blue) Taxi
  Destination and Trip Time Prediction from Partial Trajectories},'' in
  \emph{{Proceedings of the 2015th International Conference on ECML PKDD
  Discovery Challenge}}, 2015, pp. 63--74.

\bibitem{Froehlich.2008}
J.~Froehlich and J.~Krumm, ``{Route Prediction from Trip Observations},'' in
  \emph{{Society of Automotive Engineers World Congress}}, 2008.

\bibitem{DantasNobreNeto.2016}
F.~{Dantas Nobre Neto}, C.~d.~S. Baptista, and C.~E.~C. Campelo, ``{A
  user-personalized model for real time destination and route prediction},'' in
  \emph{{2016 IEEE 19th International Conference on Intelligent Transportation
  Systems (ITSC)}}, 2016, pp. 401--407.

\bibitem{Besse.2017}
P.~C. Besse, B.~Guillouet, J.-M. Loubes, and F.~Royer, ``{Destination
  Prediction by Trajectory Distribution-Based Model},'' \emph{{IEEE
  Transactions on Intelligent Transportation Systems}}, pp. 1--12, 2017.

\bibitem{Piorkowski.2009}
M.~Piorkowski, N.~Sarafijanovic-Djukic, and M.~Grossglauser, ``{CRAWDAD dataset
  epfl/mobility (v. 2009-02-24)},'' 2009.

\bibitem{Bengio.2003}
Y.~Bengio, R.~Ducharme, P.~Vincent, and C.~Janvin, ``{A Neural Probabilistic
  Language Model},'' \emph{{J. Mach. Learn. Res.}}, vol.~3, pp. 1137--1155,
  2003.

\bibitem{Xu.2015}
K.~Xu, J.~Ba, R.~Kiros, K.~Cho, A.~Courville, R.~Salakhudinov, R.~Zemel, and
  Y.~Bengio, ``{Show, Attend and Tell: Neural Image Caption Generation with
  Visual Attention},'' in \emph{{Proceedings of the 32nd International
  Conference on Machine Learning}}, 2015, pp. 2048--2057.

\end{thebibliography}
\end{document}